\documentclass[10pt,preprint]{article}
\usepackage{rlc}

\usepackage{booktabs}

\usepackage[english]{babel}

\usepackage[letterpaper,top=2cm,bottom=2cm,left=3cm,right=3cm,marginparwidth=1.75cm]{geometry}
\usepackage{algorithm}
\usepackage{algpseudocode}
\usepackage{amsmath}
\usepackage{graphicx}
\usepackage{natbib}
\usepackage[colorlinks=true, allcolors=blue]{hyperref}
\usepackage{etoolbox}
\pretocmd{\section}{\phantomsection}{}{}
\pretocmd{\subsection}{\phantomsection}{}{}
\usepackage{tikz}
\usepackage{tcolorbox}
\usepackage{xcolor}
\usetikzlibrary{shapes,arrows,positioning,decorations.pathmorphing}

\usepackage{authblk} 
\usepackage{amssymb}            
\usepackage{mathtools}          
\usepackage{mathrsfs}           
\usepackage{amsthm}

\mathtoolsset{showonlyrefs}     
\usepackage{graphicx}           
\usepackage{subcaption}         
\usepackage[space]{grffile}     
\usepackage{url}                
\usepackage{multirow} 
\title{\vspace{+1cm}EnvX: Agentize Everything with Agentic AI 
}

\author{
  \name{Linyao Chen}$^{1,4,\ast}$ \quad
  \name{Zimian Peng}$^{1,2,5,\ast}$ \quad
  \name{Yingxuan Yang}$^{3}$ \\
  \vspace{-0.3cm}
  \name{Yikun Wang}$^{1,2,6}$ \quad
  \name{Wenzheng Tom Tang}$^{1}$ \\
  \name{Hiroki H. Kobayashi}$^{4}$ \quad
  \name{Weinan Zhang}$^{2,3,^\dag}$ \\
  \vspace{+0.5cm} 
  $^{1}$EnvX Team \quad
  $^{2}$Shanghai Innovation Institute \quad
  $^{3}$Shanghai Jiao Tong University \\
  $^{4}$The University of Tokyo \quad
  $^{5}$Zhejiang University \quad
  $^{6}$Fudan University \\
  $^{\ast}$Equal contribution \quad
  $^{\dag}$Corresponding author.
}

\usepackage{graphicx}
\usepackage{fancyhdr}

\begin{document}
\thispagestyle{fancy} 

\maketitle{}

\pagestyle{plain}

\begin{abstract}
The widespread availability of open-source repositories has led to a vast collection of reusable software components, yet their utilization remains manual, error-prone, and disconnected. Developers must navigate documentation, understand APIs, and write integration code, creating significant barriers to efficient software reuse. To address this, we present EnvX, a framework that leverages Agentic AI to agentize GitHub repositories, transforming them into intelligent, autonomous agents capable of natural language interaction and inter-agent collaboration. Unlike existing approaches that treat repositories as static code resources, EnvX reimagines them as active agents through a three-phase process: (1) TODO-guided environment initialization, which sets up the necessary dependencies, data, and validation datasets; (2) human-aligned agentic automation, allowing repository-specific agents to autonomously perform real-world tasks; and (3) Agent-to-Agent (A2A) protocol, enabling multiple agents to collaborate. By combining large language model capabilities with structured tool integration, EnvX automates not just code generation, but the entire process of understanding, initializing, and operationalizing repository functionality. We evaluate EnvX on the GitTaskBench benchmark, using 18 repositories across domains such as image processing, speech recognition, document analysis, and video manipulation. Our results show that EnvX achieves a 74.07\% execution completion rate and 51.85\% task pass rate, outperforming existing frameworks. Case studies further demonstrate EnvX's ability to enable multi-repository collaboration via the A2A protocol. This work marks a shift from treating repositories as passive code resources to intelligent, interactive agents, fostering greater accessibility and collaboration within the open-source ecosystem.

\textbf{Keywords:} Agentization, Agentic AI, Code Repository, Multi-Agent System
\end{abstract}

\section{Introduction}\label{sec:intro}

In the field of artificial intelligence, an agent is an autonomous system capable of perceiving its environment, reasoning about goals, and taking actions to achieve them. Agentization~\citep{acharya2025agentic, sapkota2025ai, yang2025agenticwebweavingweb} refers to the process of transforming various entities, such as environments, code, and services, into agents that not only retain their original functionalities but also gain the capacity for autonomous action and communication. 
This concept lays the foundation for EnvX, a novel framework designed to "agentize everything" in the software ecosystem. Through the power of agentic AI, EnvX transforms open-source repositories into intelligent, interactive agents that not only retain their core functionalities but can also autonomously respond to natural language instructions, collaborate with other agents, and adapt to changing requirements.

The proliferation of open-source software repositories has fundamentally transformed modern software development, with platforms like GitHub hosting millions of projects that serve as building blocks for countless applications. However, the current paradigm of repository utilization~\citep{asadi201repoutilization,ma2025alibaba} remains fundamentally manual: developers must navigate documentation, understand APIs, examine code examples, and write integration code before they can leverage a repository's functionality. This process is not only time-consuming but also error-prone, creating significant barriers to efficient software reuse and collaboration. Recent advances in Large Language Models (LLMs) have sparked efforts to automate this process~\citep{chen2025repoforgetrainingsotafastthinking,luo2024repoagentllmpoweredopensourceframework,wang2025repomasterautonomousexplorationunderstanding}, with systems demonstrating capabilities in repository understanding, code generation, and automated task execution. However, these approaches primarily treat repositories as sources for code generation or targets for modification, rather than transforming them into interactive agents that users can directly interact with through natural language. The challenge of enabling seamless natural language communication with repository functionalities remains largely unaddressed.

This paper introduces EnvX, a novel framework that automatically transforms arbitrary GitHub repositories into agents with agentic automation and agentic communication capacity. Unlike traditional NL2Code approaches~\citep{zan2023NL2code} that focus on generating new code, EnvX enables users to directly invoke existing repository functionalities through natural language instructions. More importantly, EnvX extends beyond single-agent interactions by implementing the Agent-to-Agent (A2A) protocol, enabling multiple repository agents to collaborate and form an ecosystem of intelligent software components.
The agentization of arbitrary repositories presents several fundamental challenges. Repositories exhibit extreme heterogeneity in structure, dependencies, and usage patterns. Understanding repository semantics transcends mere code parsing, which means the framework must comprehend intended functionality, identify entry points, and infer usage patterns from often incomplete documentation. Additionally, enabling meaningful inter-agent communication requires establishing common protocols while preserving repository-specific capabilities.

Our key insight is that by combining LLM capabilities with structured tool integration, we can automate not just code generation but the entire process of understanding, initializing, and operationalizing repository functionality. This shift from treating repositories as passive code resources to reimagining them as active, intelligent agents represents a fundamental paradigm change in how developers discover, understand, and utilize software components.
We evaluate EnvX on GitTaskBench, demonstrating its effectiveness across diverse domains including image processing, speech recognition, document analysis, and video manipulation. Our results show that EnvX significantly outperforms existing code agent frameworks, validating our hypothesis that systematic agentization can make complex repositories accessible through natural language interaction.

The remainder of this paper is organized as follows: Section 2 reviews related work. Section 3 presents the EnvX framework and our three-phase agentization process. Section 4 describes our experimental setup and results. Section 5 discusses implications and limitations. Section 6 concludes with future directions.

\section{Related Work}
\subsection{Language Models as Agents}
The integration of Large Language Models (LLMs) ~\citep{deepseekai2025deepseekv3, openai2024gpt4o} as autonomous agents ~\citep{li2023camel, huang2024llmagents} has emerged as a transformative paradigm, enabling them to plan, reason, and execute complex tasks. ReAct~\citep{yao2023reactsynergizingreasoningacting} exemplifies this by combining reasoning and acting, allowing LLMs to generate both reasoning traces and task-specific actions for autonomous problem solving. Parallel efforts have explored multi-agent systems (MAS)~\citep{Yang2024LLMbasedMS, zhang2025avengers}, where multiple LLM agents collaborate on complex tasks. In software engineering, ChatDev~\citep{qian2024chatdevcommunicativeagentssoftware} demonstrates this potential by simulating a software company with role-based agents, while MetaGPT~\citep{hong2024metagpt} extends the approach with standardized operating procedures inspired by human practices. These advances pave the way for structured, collaborative interactions in software development.
What differentiates our work is the introduction of an Agent-to-Agent (A2A) protocol~\citep{a2a2025, yang2025surveyaiagentprotocols}, which enables multiple repository agents to communicate, collaborate, and coordinate on tasks, thereby forming a cohesive ecosystem of intelligent components capable of addressing complex challenges.

\subsection{Software Engineering for Repository}
Another important line of work focuses on software repositories themselves, moving beyond their role as static code collections toward active, intelligent entities. RepoAgent~\citep{luo2024repoagentllmpoweredopensourceframework} leverages LLMs to automatically generate and maintain repository documentation, improving developer comprehension and usability of complex codebases. Building on this idea, RepoMaster~\citep{wang2025repomasterautonomousexplorationunderstanding} autonomously explores and analyzes GitHub repositories, including structures and dependencies, to enable context-aware code execution and task solving, achieving strong performance on benchmarks such as GitTaskBench~\citep{ni2025gittaskbench} and OpenAct~\citep{lyu-etal-2025-openact}. From a data and training perspective, RepoForge~\citep{chen2025repoforgetrainingsotafastthinking} complements these efforts by constructing large-scale executable environments and providing efficient labeling pipelines for agent training in software engineering.
Despite these advances in repository intelligence, most existing systems still focus primarily on code modification and issue resolution rather than enabling natural language interaction. SWE-Agent~\citep{yang2024sweagent} equips LLMs with a specialized interface for navigating and editing codebases, OpenHands~\citep{wang2025openhands} provides a flexible platform for building agents that interact with software environments, and Aider~\citep{aider2025} delivers a terminal-based assistant tightly integrated with repositories. While these approaches excel at automated code editing and bug fixing, they treat repositories as targets for modification rather than as interactive partners. They still demand significant manual effort to fully understand repository functionality and fall short of transforming repositories into agents capable of natural language interaction.

\section{Methodology}
In this section, we present \textbf{EnvX}, a novel LLM-based framework designed to transcend a purely tool-calling workflow by incorporating carefully engineered tools that enable the systematic agentization of arbitrary repositories. Distinct from prior approaches, EnvX seamlessly integrates specialized tools into the agentization pipeline, thereby transforming the process into one that is both reliable and efficient. Specifically, Section~3.1 delineates the overall system workflow, while Section~3.2 elaborates on the design and integration of the key tools.


\begin{figure}[t]
    \centering
    \begin{subfigure}{0.6\linewidth}
        \centering
        \includegraphics[width=\linewidth]{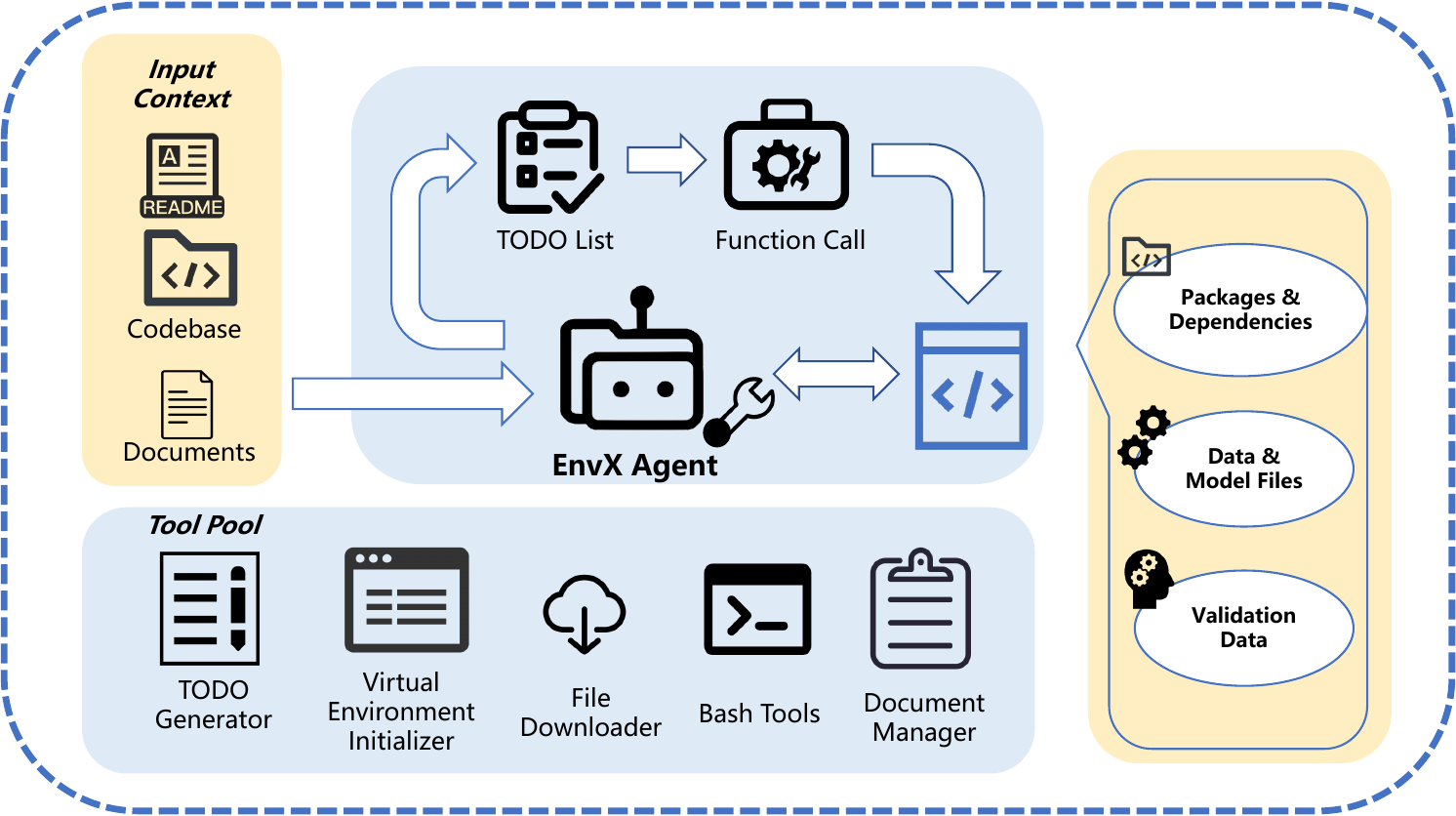}
        \caption{Phase 1: Agentic Environment Setting}
    \end{subfigure}
    
    \vspace{1em} 
    
    \begin{subfigure}{\linewidth}
        \centering
        \includegraphics[width=\linewidth]{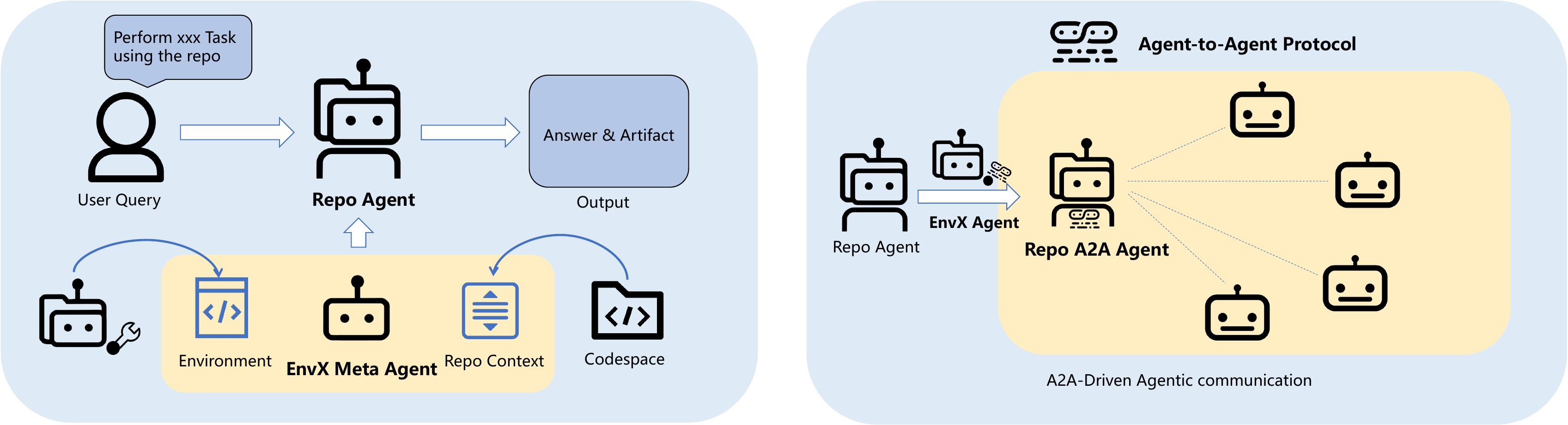}
        \caption{Phase 2: Agentic Automation \& Phase 3: Agentic Communication}
    \end{subfigure}
    
    \caption{Overview of EnvX workflow. Three phases are defined for the operation, including the Agentic Environment Setting phase, Agentic Automation phase for solving real-world user queries, and Agentic Communication Phase to introduce generated repository agent to multi-agent communication and collaboration.}
    \label{fig:envx-overview}
    
\end{figure}

\subsection{System Overview}

As illustrated in Fig.~\ref{fig:envx-overview}, the agentization of a repository within EnvX proceeds through three progressive phases.  
In the first phase, the system establishes the computational environment necessary for interacting with the target repository. This requires the agent to acquire a comprehensive understanding of the repository’s codebase and its accompanying documentation (e.g., README files). During this stage, the agent systematically explores the repository and generates a structured representation of its context to support subsequent phases.  

In the second phase, EnvX instantiates a repository-specific agent capable of addressing user queries and tasks in a manner that emulates human reasoning and operations. The repository agent is derived by integrating the environment constructed in Phase~1 with the extracted repository context, thereby enabling task execution through agentic automation.  

In the third phase, EnvX augments the repository agent with communication capabilities based on the Agent-to-Agent (A2A) protocol. This protocol establishes a standardized schema for inter-agent communication through the generation of agent cards and the extraction of agent skills. Consequently, the repository agent acquires the ability to interact, collaborate, and coordinate with other agents, thereby facilitating multi-agent cooperation and enhancing system-level scalability and robustness.  

\subsection{TODO-Guided Environment Initialization}

As depicted in Fig.~\ref{fig:envx-overview}, Phase~1 focuses on constructing a comprehensive environment that empowers the agent to operationalize the functionalities of the original repository. Unlike traditional approaches that narrowly define the environment as a collection of packages and dependencies, EnvX conceptualizes the environment as comprising three critical components: (1) dependencies and packages, (2) data and model artifacts, and (3) validation datasets. This broader definition ensures that the environment can robustly support repository operations while providing verifiable reliability through validation data.  

A central mechanism in this phase is the generation of a structured TODO list. Based on an in-depth analysis of guidance documents (e.g., README files and technical manuals), EnvX automatically produces a comprehensive set of initialization tasks. A dedicated TODO management tool subsequently maintains and executes these tasks, orchestrating the entire environment initialization workflow. Moreover, the system iteratively revises the TODO list in response to execution errors, thereby improving adaptability. By embedding TODO-driven control into initialization, EnvX not only systematizes the workflow and enhances goal clarity and decomposition, but also strengthens monitoring and self-reflection within the agentic system.  

\subsection{Human-Aligned Agentic Automation}

During Phase~2, EnvX constructs repository agents that autonomously perform real-world tasks through tool-mediated automation. To ensure alignment with human usage scenarios, we design a meta-agent framework equipped with widely adopted development and management tools. By leveraging the environment initialized in Phase~1 and the repository context extracted from the codebase, the meta-agent is agentized as a repository-specific agent. Compared with meta-agent, the agentized repository agent understands and integrates the functionalities in the original repository, with the capacity to call the functionalities to solve real-world tasks.

These agents are capable of receiving diverse user queries and tasks and producing responses in a manner that aligns with human expectations of utility, reliability, and interpretability. When the users interact with system via the input queries, the repository agent tries to reason over given queries, and invokes tools and repository functionalities to complete the requirements of the input queries.  

\subsection{Agentization for Agentic Communication}

Research on multi-agent communication~\citep{sukhbaatar2016commnet,zhang2025gdesigner} has garnered substantial scholarly interest, as it enables distributed systems to achieve coordination, cooperation, and collective problem-solving beyond the capacity of individual agents. Moreover, effective communication mechanisms are essential for ensuring the scalability, robustness, and adaptability of multi-agent systems. In Phase~3, EnvX equips the repository agent with communication competencies through the A2A protocol. This protocol prescribes a standardized communication interface, operationalized via the construction of agent cards and the formalization of agent skills using a predefined A2A toolbox. By implementing an A2A communication port grounded in the repository context and initialized environment, EnvX facilitates seamless inter-agent collaboration and enables higher-order forms of agentic intelligence.  

\subsection{Agentic Tool Integration}

Section~3.1 outlined the overall agentization pipeline, which operates through a pure tool-calling paradigm. Within this paradigm, tools assume a pivotal role in enabling the diverse functionalities required by the agent. We categorize the tools into six principal classes:  

\begin{itemize}  
    \item \textbf{Basic Tools}: Fundamental utilities that support core agent functions such as reasoning, file reading and writing, script execution, and task completion. These tools also serve as the foundation of the meta-agent, ensuring that the repository agent achieves baseline functionality.  
    \item \textbf{File Download Tool}: A utility that enables agents to acquire necessary files, particularly during the environment setup phase, thereby facilitating the preparation of datasets and model checkpoints required for downstream tasks.  
    \item \textbf{TODO Management Tool}: Central to Phase~1, this tool provides three essential capabilities: (1) initialization of structured TODO lists, (2) automatic generation of validation data for new tasks, and (3) systematic verification of task completion. By integrating these capabilities, the tool ensures that initialization workflows are reliable, traceable, and aligned with the repository’s requirements.  
    \item \textbf{Dependency Management Tool}: A unified mechanism for handling repository-specific dependencies. Given that repositories may adopt heterogeneous initialization practices (e.g., \texttt{requirements.txt}, Conda environments), this tool abstracts and manages dependency installation to guarantee a consistent execution environment.  
    \item \textbf{Code Knowledge Graph Tool}: A semantic analysis component that extracts the repository’s primary functionalities and potential applications. It enables the construction of a Code Knowledge Graph (CKG), which agents can query during task automation to support reasoning and enhance interpretability.  
    \item \textbf{A2A Generation Tool}: A specialized module that endows repository agents with agentic communication abilities. Following human-like operational logic, the tool identifies agent skills, initializes services, generates corresponding implementation code, and produces agent cards, thereby enabling seamless integration with other agents through the A2A protocol.  
\end{itemize}  

\section{Empirical Study}
In this section, we report the preliminary experimental results of EnvX on open benchmarks and provide case studies.

\subsection{Benchmarks and Metrics}
To verify the effectiveness of our system, we adopt GitTaskBench~\citep{ni2025gittaskbench} as our evaluation benchmark. 
GitTaskBench targets open-source repositories, thereby reflecting practical, real-world tasks. 
The benchmark comprises 18 repositories spanning multiple domains (image, speech, document, video, and others), enabling assessment of agent capabilities under complex, realistic scenarios. 
It further provides 54 human-validated tasks and a rigorous evaluation pipeline to ensure trustworthy assessment.

Following the design of GitTaskBench, we employ the following metrics:
\begin{itemize}
    \item \textbf{Execution Completion Rate (ECR)}: Measures whether the repository can be executed to produce evaluable outputs by inspecting the generated files. Specifically, it verifies that the expected output files exist, are non-empty, and can be processed by the official evaluation scripts.  
    \item \textbf{Task Pass Rate (TPR)}: Computed on top of ECR to quantify task-level performance quality. GitTaskBench provides domain-specific evaluation scripts that compare the generated outputs with ground truth using appropriate criteria for each domain. This metric reflects the overall task success rate of a repository agent.
    \item \textbf{Token Costs}: Quantifies efficiency in terms of LLM usage. We report both input and output tokens.
\end{itemize}

\subsection{Evaluation Setup}
We compare our system against three existing coding-agent frameworks designed for repository-level task execution, using three widely adopted LLM backbones. Details are as follows.

\subsubsection{Baselines}
As described above, we implement the following baselines for comparison:
\begin{itemize}
    \item \textbf{OpenHands}~\citep{wang2025openhands}: An LLM-driven coding-agent architecture that provides sandboxed environments for code execution and API interactions. We deploy the repositories in local sandboxes and evaluate each query in its corresponding environment.
    \item \textbf{Aider}~\citep{aider2025}: A terminal-based AI coding agent that operates on top of existing repositories/codebases. We execute each task within the corresponding repository following Aider’s standard workflow.
    \item \textbf{SWE-Agent}~\citep{yang2024sweagent}: An LLM-driven agent for software engineering, widely used for autonomous GitHub repository understanding and issue resolution. We evaluate it using its default workflow to assess its agentic capabilities. 
\end{itemize}
We use three widely adopted LLMs: GPT-4o~\citep{openai2024gpt4o}, GPT-4.1~\citep{openai2024gpt41}, and Claude 3.7~\citep{claude2024} Sonnet as the backbone models for both our system and the baselines, enabling a comprehensive and fair comparison.

\subsubsection{Implementation Details}
In our experiments, we implement the EnvX system with the tools described in Section~3.1, issuing one function call per round to complete environment setup, agentic automation, and agent communication. We cap the maximum number of steps at 200 and allow parallel tool calls within the workflow. To mitigate potential issues caused by network or system instability, we set the maximum number of retries to 10. 
Because the original benchmark evaluation scripts were not directly compatible with our system, we configured EnvX to write outputs to the designated directories; we then determine execution success and task success by verifying the outputs and comparing them against the ground truth.

For the baselines, we follow the official experimental settings and the reporting result of GitTaskBench~\citep{ni2025gittaskbench}.

\begin{table*}[t]
\centering
\resizebox{\textwidth}{!}{
\begin{tabular}{l l c c c c c}
\toprule
\textbf{Framework} & \textbf{LLM} & \textbf{ECR (\%)} $\uparrow$ & \textbf{TPR (\%)} $\uparrow$ & \textbf{Input Tokens (k)} $\downarrow$ & \textbf{Output Tokens} $\downarrow$ \\
\midrule
\multirow{3}{*}{Aider} 
  & GPT-4o       & 5.56  & 1.85  & 10.67 & 492.67   \\
  & GPT-4.1      & 11.11 & 7.41  & 14.83 & 734.17   \\
  & Claude 3.5   & 16.67 & 12.96 & 7.48 & 534.00   \\
\hline
\multirow{3}{*}{SWE-Agent}
  & GPT-4o       & 17.58 & 10.19 & 275.53 & 1282.70  \\
  & GPT-4.1      & 38.89 & 31.48 & 301.11 & 2098.33  \\
  & Claude 3.7   & 64.81 & 42.59 & 552.79 & 807.63  \\
\hline
\multirow{3}{*}{OpenHands}
  & GPT-4o       & 21.30 & 14.82 & 760.53  & 3990.31  \\
  & GPT-4.1      & 55.56 & 42.59 & 465.94  & 1535.47   \\
  & Claude 3.7   & 72.22 & 48.15 & 9501.25 & 85033.05 \\
\hline
\multirow{3}{*}{EnvX} 
  & GPT-4o       & 42.59 & 33.33 & 811.02  & 4506.83  \\
  & GPT-4.1      & 68.52 & 46.30 & 380.20  & 1648.43  \\
  & Claude 3.7   & \textbf{74.07} & \textbf{51.85} & 562.56 &  5686.89 \\
\bottomrule
\end{tabular}}
\caption{Comparison of different LLMs across frameworks (Aider, SWE-Agent, OpenHands) with evaluation metrics: ECR, TPR, Input Tokens, Output Tokens, and Cost. Bold indicates best accuracy results.}
\label{tab:llm_comparison}
\end{table*}

\subsection{Experimental Analysis}
The main experimental results are reported in Table~\ref{tab:llm_comparison}. 
On GitTaskBench, our system achieves state-of-the-art performance for repository automation. 
With Claude 3.7 Sonnet as the backbone, our system attains a TPR exceeding 50\%, improves upon prior work by 7.6\%, and achieves the best ECR to date at 74.07\%. 
Furthermore, owing to the well-designed tool workflow, our system exhibits strong robustness across backbone models. 
On GPT-4.1, it improves ECR by 23.40 percentage points and TPR by 8.72 percentage points over the previous best. 
Moreover, with GPT-4o, we observe relative improvements of 100\% in ECR and 124.90\% in TPR, highlighting the substantial potential to enhance agentic capabilities.

On the other hand, our system is more efficient than prior work in repository automation, especially with larger-parameter models. 
Compared with OpenHands~\citep{wang2025openhands} which performs comparably yet consumes more than 10$\times$ the input and output tokens when using Claude 3.7 Sonnet. We conclude that our system is markedly more efficient while maintaining effectiveness. 
For smaller models, although our approach incurs additional tokens due to tool calls, it yields significant performance gains. 
Moreover, when comparing GPT-4o and GPT-4.1, which share the same architecture but differ in size, the token cost decreases with scale, likely because larger models plan more effectively and make fewer erroneous steps. 
This suggests considerable headroom for efficient agentic automation when stronger backbones are incorporated.

\subsection{Case Study}

\begin{figure}[t]
    \centering
    \includegraphics[width=\linewidth]{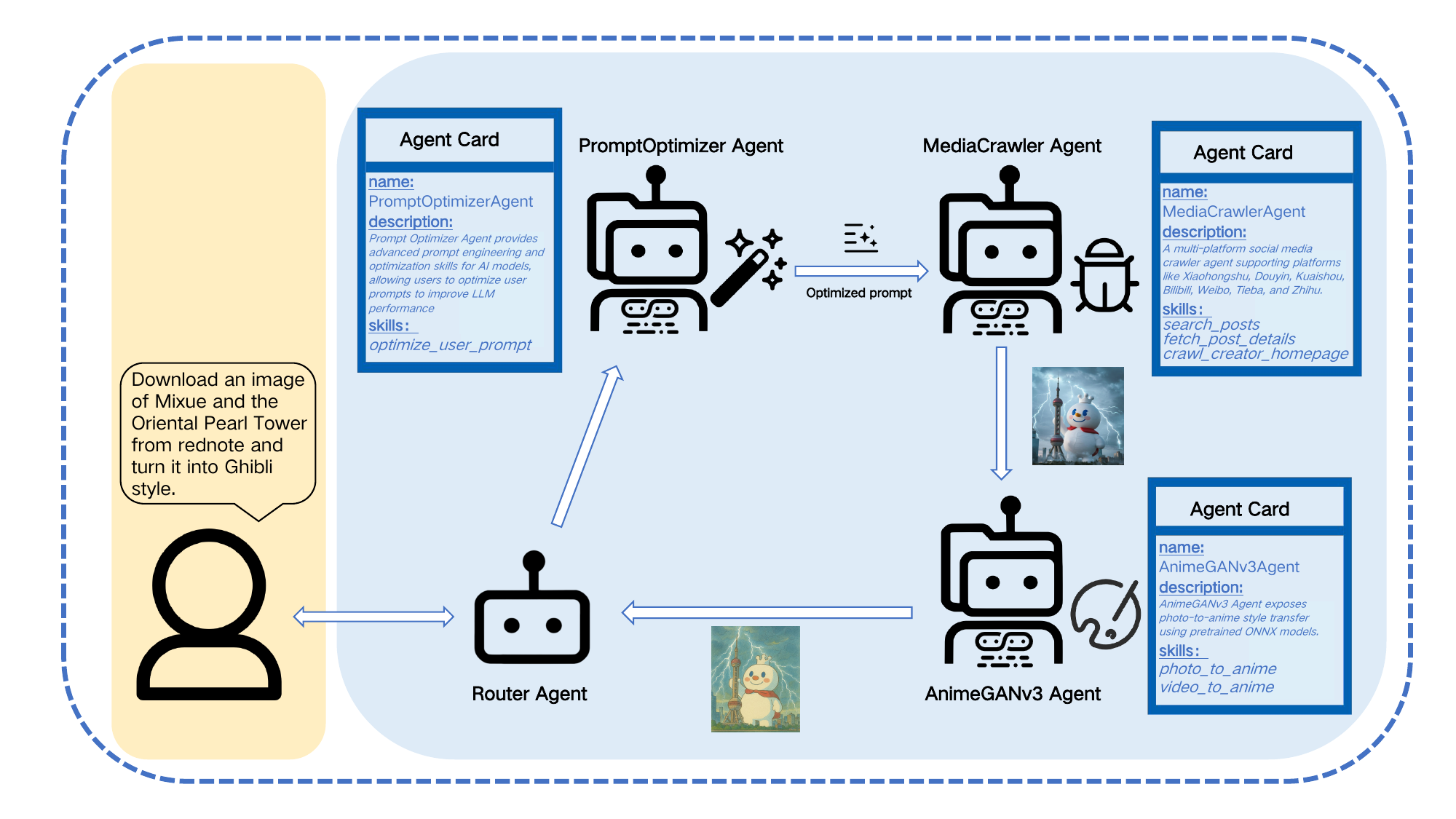}
    \caption{Case study on repository agents collaboration. Three repositories are agentized by EnvX. Router agents in A2A system obtain the agent skills via agent cards, solve complex real-world tasks under the collaboration between repository agents.}
    \label{fig:a2a}
\end{figure}

In this section, we demonstrate the potential of repository agents in addressing real-world tasks through agentic communication. As illustrated in Fig.~\ref{fig:a2a}, when tackling complex tasks that depend on multiple repositories, the system first transforms each repository into a repository agent and generates corresponding agent cards containing domain knowledge and invocation methods. The router agent within the A2A system then coordinates these repository agents, invoking them individually and integrating their functionalities into a coherent workflow. This workflow converts an input prompt into an output figure with specialized styles. This case study highlights the reliability of the agentization process enabled by EnvX and underscores its immense potential to harness the value of the open-source ecosystem for real-world applications.

\section{Discussion}
EnvX demonstrates that \emph{agentization}, which transforms heterogeneous repositories into interactive agents and coordinates them via the Agent-to-Agent (A2A) protocol, can reliably address practical tasks. Nevertheless, several limitations remain. First, our current evaluation primarily relies on scripted oracles and curated tasks, which leaves important gaps in coverage for long-horizon coordination, robustness under distribution shift, and security-in-the-loop failure modes. Second, while we have validated \emph{hundreds} of A2A interactions in practice, the verification signals are still coarse-grained at times and thus constrain the automatic synthesis and selection of high-quality A2A agents. Looking ahead, we will (i) scale A2A validation and make it more \emph{reliable} by systematically generating richer verification data and oracles---combining input--output pairs, property-based checks, and metamorphic relations---to provide precise, reproducible pass/fail signals for agent synthesis; (ii) strengthen the standardization of agent cards and skill schemas with explicit contracts, versioning, and provenance logging to support safe reuse; and (iii) study cost--quality trade-offs across data, tools, and model backbones to guide principled scaling of agentization. We believe these directions will transform A2A from a validated prototype into a dependable foundation for building, verifying, and maintaining large ecosystems of repository agents, thereby maximizing the potential of the open-source ecosystem.

\section{Conclusion}
Overall, this research introduces EnvX, a novel agent system that transforms raw open-source repositories into intelligent repository agents capable of providing comprehensive automation and communication services. We demonstrate the effectiveness and robustness of our approach through extensive experiments and real-world case studies, highlighting how each component of the methodology contributes to successful agentization. Beyond performance improvements, EnvX fosters a sustainable ecosystem of collaborative agents by converting existing repositories into communicative agents and orchestrating their interactions to address complex real-world problems. We believe that agentizing repositories in this way will inspire future research on multi-agent collaboration and further advance the open-source community.

\bibliographystyle{plainnat}
\bibliography{reference}

\begin{thebibliography}{29}
\providecommand{\natexlab}[1]{#1}
\providecommand{\url}[1]{\texttt{#1}}
\expandafter\ifx\csname urlstyle\endcsname\relax
  \providecommand{\doi}[1]{doi: #1}\else
  \providecommand{\doi}{doi: \begingroup \urlstyle{rm}\Url}\fi

\bibitem[aid()]{aider2025}
Aider: Ai pair programming in your terminal.
\newblock \url{https://aider.chat/}.
\newblock Accessed: 2025-09-06.

\bibitem[Acharya et~al.(2025)Acharya, Kuppan, and Divya]{acharya2025agentic}
Deepak~Bhaskar Acharya, Karthigeyan Kuppan, and B~Divya.
\newblock Agentic ai: Autonomous intelligence for complex goals--a comprehensive survey.
\newblock \emph{IEEe Access}, 2025.

\bibitem[{Anthropic}(2024)]{claude2024}
{Anthropic}.
\newblock Claude sonnet 3.7.
\newblock \url{https://www.anthropic.com}, 2024.
\newblock Large language model.

\bibitem[Asadi et~al.(2019)Asadi, Abdullah, Yah, and Nazir]{asadi201repoutilization}
Shahla Asadi, Rusli Abdullah, Yusmadi Yah, and Shah Nazir.
\newblock Understanding institutional repository in higher learning institutions: A systematic literature review and directions for future research.
\newblock \emph{IEEE Access}, 7:\penalty0 35242--35263, 2019.

\bibitem[Chen et~al.(2025)Chen, Zhao, Chen, Lin, Chen, Leung, Rajbahadur, Oliva, Zhang, Bhatia, Yong, and Hassan]{chen2025repoforgetrainingsotafastthinking}
Zhilong Chen, Chengzong Zhao, Boyuan Chen, Dayi Lin, Yihao Chen, Arthur Leung, Gopi~Krishnan Rajbahadur, Gustavo~A. Oliva, Haoxiang Zhang, Aaditya Bhatia, Chong~Chun Yong, and Ahmed~E. Hassan.
\newblock Repoforge: Training a sota fast-thinking swe agent with an end-to-end data curation pipeline synergizing sft and rl at scale, 2025.
\newblock URL \url{https://arxiv.org/abs/2508.01550}.

\bibitem[DeepSeek-AI et~al.(2025)DeepSeek-AI, Liu, Feng, Pan, et~al.]{deepseekai2025deepseekv3}
DeepSeek-AI, Aixin Liu, Bei Feng, Zizheng Pan, et~al.
\newblock Deepseek-v3 technical report, 2025.
\newblock URL \url{https://arxiv.org/abs/2412.19437}.

\bibitem[Google(2025)]{a2a2025}
Google.
\newblock A2a: Agent2agent protocol, 2025.
\newblock URL \url{https://github.com/google/A2A}.
\newblock Accessed: 2025-04-21.

\bibitem[Hong et~al.(2024)Hong, Zhuge, Chen, Zheng, Cheng, Wang, Zhang, Wang, Yau, Lin, Zhou, Ran, Xiao, Wu, and Schmidhuber]{hong2024metagpt}
Sirui Hong, Mingchen Zhuge, Jonathan Chen, Xiawu Zheng, Yuheng Cheng, Jinlin Wang, Ceyao Zhang, Zili Wang, Steven Ka~Shing Yau, Zijuan Lin, Liyang Zhou, Chenyu Ran, Lingfeng Xiao, Chenglin Wu, and J{\"u}rgen Schmidhuber.
\newblock Meta{GPT}: Meta programming for a multi-agent collaborative framework.
\newblock In \emph{The Twelfth International Conference on Learning Representations}, 2024.
\newblock URL \url{https://openreview.net/forum?id=VtmBAGCN7o}.

\bibitem[Huang et~al.(2024)Huang, Liu, Chen, Wang, Wang, Lian, Wang, Tang, and Chen]{huang2024llmagents}
Xu~Huang, Weiwen Liu, Xiaolong Chen, Xingmei Wang, Hao Wang, Defu Lian, Yasheng Wang, Ruiming Tang, and Enhong Chen.
\newblock Understanding the planning of llm agents: A survey, 2024.
\newblock URL \url{https://arxiv.org/abs/2402.02716}.

\bibitem[Li et~al.(2023)Li, Hammoud, Itani, Khizbullin, and Ghanem]{li2023camel}
Guohao Li, Hasan Abed Al~Kader Hammoud, Hani Itani, Dmitrii Khizbullin, and Bernard Ghanem.
\newblock Camel: Communicative agents for "mind" exploration of large language model society, 2023.
\newblock URL \url{https://arxiv.org/abs/2303.17760}.

\bibitem[Luo et~al.(2024)Luo, Ye, Liang, Zhang, Qin, Lu, Wu, Cong, Lin, Zhang, Che, Liu, and Sun]{luo2024repoagentllmpoweredopensourceframework}
Qinyu Luo, Yining Ye, Shihao Liang, Zhong Zhang, Yujia Qin, Yaxi Lu, Yesai Wu, Xin Cong, Yankai Lin, Yingli Zhang, Xiaoyin Che, Zhiyuan Liu, and Maosong Sun.
\newblock Repoagent: An llm-powered open-source framework for repository-level code documentation generation, 2024.
\newblock URL \url{https://arxiv.org/abs/2402.16667}.

\bibitem[Lyu et~al.(2025)Lyu, Cong, Yu, Yang, Qian, Wang, Qin, Ye, Lu, Qian, Zhang, Yan, Lin, Liu, and Sun]{lyu-etal-2025-openact}
Bohan Lyu, Xin Cong, Heyang Yu, Pan Yang, Cheng Qian, Zihe Wang, Yujia Qin, Yining Ye, Yaxi Lu, Chen Qian, Zhong Zhang, Yukun Yan, Yankai Lin, Zhiyuan Liu, and Maosong Sun.
\newblock Enhancing open-domain task-solving capability of {LLM}s via autonomous tool integration from {G}it{H}ub.
\newblock In Wanxiang Che, Joyce Nabende, Ekaterina Shutova, and Mohammad~Taher Pilehvar, editors, \emph{Proceedings of the 63rd Annual Meeting of the Association for Computational Linguistics (Volume 1: Long Papers)}, pages 17257--17277, Vienna, Austria, July 2025. Association for Computational Linguistics.
\newblock ISBN 979-8-89176-251-0.
\newblock \doi{10.18653/v1/2025.acl-long.845}.
\newblock URL \url{https://aclanthology.org/2025.acl-long.845/}.

\bibitem[Ma et~al.(2025)Ma, Yang, Cao, Li, Huang, and Li]{ma2025alibaba}
Yingwei Ma, Qingping Yang, Rongyu Cao, Binhua Li, Fei Huang, and Yongbin Li.
\newblock Alibaba lingmaagent: Improving automated issue resolution via comprehensive repository exploration.
\newblock In \emph{Proceedings of the 33rd ACM International Conference on the Foundations of Software Engineering}, pages 238--249, 2025.

\bibitem[Ni et~al.(2025)Ni, Wang, Zhang, Lu, He, You, Tang, Du, Sun, Liu, Hu, Chen, Li, Li, Hu, Jiao, Jiang, and Lyu]{ni2025gittaskbench}
Ziyi Ni, Huacan Wang, Shuo Zhang, Shuo Lu, Ziyang He, Wang You, Zhenheng Tang, Yuntao Du, Bill Sun, Hongzhang Liu, Sen Hu, Ronghao Chen, Bo~Li, Xin Li, Chen Hu, Binxing Jiao, Daxin Jiang, and Pin Lyu.
\newblock Gittaskbench: A benchmark for code agents solving real-world tasks through code repository leveraging, 2025.
\newblock URL \url{https://arxiv.org/abs/2508.18993}.

\bibitem[OpenAI(2024)]{openai2024gpt41}
OpenAI.
\newblock Gpt-4.1 technical report.
\newblock \url{https://openai.com/research/gpt-4-1}, 2024.
\newblock Accessed: 2025-09-09.

\bibitem[OpenAI et~al.(2024)OpenAI, :, Hurst, Lerer, Goucher, Zhang, Jin, Dai, Malkov, et~al.]{openai2024gpt4o}
OpenAI, :, Aaron Hurst, Adam Lerer, Adam~P. Goucher, Yuchen Zhang, Yujia Jin, Yunxing Dai, Yury Malkov, et~al.
\newblock Gpt-4o system card, 2024.
\newblock URL \url{https://arxiv.org/abs/2410.21276}.

\bibitem[Qian et~al.(2024)Qian, Liu, Liu, Chen, Dang, Li, Yang, Chen, Su, Cong, Xu, Li, Liu, and Sun]{qian2024chatdevcommunicativeagentssoftware}
Chen Qian, Wei Liu, Hongzhang Liu, Nuo Chen, Yufan Dang, Jiahao Li, Cheng Yang, Weize Chen, Yusheng Su, Xin Cong, Juyuan Xu, Dahai Li, Zhiyuan Liu, and Maosong Sun.
\newblock Chatdev: Communicative agents for software development, 2024.
\newblock URL \url{https://arxiv.org/abs/2307.07924}.

\bibitem[Sapkota et~al.(2025)Sapkota, Roumeliotis, and Karkee]{sapkota2025ai}
Ranjan Sapkota, Konstantinos~I Roumeliotis, and Manoj Karkee.
\newblock Ai agents vs. agentic ai: A conceptual taxonomy, applications and challenges.
\newblock \emph{arXiv preprint arXiv:2505.10468}, 2025.

\bibitem[Sukhbaatar et~al.(2016)Sukhbaatar, Szlam, and Fergus]{sukhbaatar2016commnet}
Sainbayar Sukhbaatar, Arthur Szlam, and Rob Fergus.
\newblock Learning multiagent communication with backpropagation, 2016.
\newblock URL \url{https://arxiv.org/abs/1605.07736}.

\bibitem[Wang et~al.(2025{\natexlab{a}})Wang, Ni, Zhang, Lu, Hu, He, Hu, Lin, Guo, Chen, Li, Jiang, Du, and Lyu]{wang2025repomasterautonomousexplorationunderstanding}
Huacan Wang, Ziyi Ni, Shuo Zhang, Shuo Lu, Sen Hu, Ziyang He, Chen Hu, Jiaye Lin, Yifu Guo, Ronghao Chen, Xin Li, Daxin Jiang, Yuntao Du, and Pin Lyu.
\newblock Repomaster: Autonomous exploration and understanding of github repositories for complex task solving, 2025{\natexlab{a}}.
\newblock URL \url{https://arxiv.org/abs/2505.21577}.

\bibitem[Wang et~al.(2025{\natexlab{b}})Wang, Li, Song, Xu, Tang, Zhuge, Pan, Song, Li, Singh, Tran, Li, Ma, Zheng, Qian, Shao, Muennighoff, Zhang, Hui, Lin, Brennan, Peng, Ji, and Neubig]{wang2025openhands}
Xingyao Wang, Boxuan Li, Yufan Song, Frank~F. Xu, Xiangru Tang, Mingchen Zhuge, Jiayi Pan, Yueqi Song, Bowen Li, Jaskirat Singh, Hoang~H. Tran, Fuqiang Li, Ren Ma, Mingzhang Zheng, Bill Qian, Yanjun Shao, Niklas Muennighoff, Yizhe Zhang, Binyuan Hui, Junyang Lin, Robert Brennan, Hao Peng, Heng Ji, and Graham Neubig.
\newblock Openhands: An open platform for ai software developers as generalist agents, 2025{\natexlab{b}}.
\newblock URL \url{https://arxiv.org/abs/2407.16741}.

\bibitem[Yang et~al.(2024{\natexlab{a}})Yang, Jimenez, Wettig, Lieret, Yao, Narasimhan, and Press]{yang2024sweagent}
John Yang, Carlos~E. Jimenez, Alexander Wettig, Kilian Lieret, Shunyu Yao, Karthik Narasimhan, and Ofir Press.
\newblock Swe-agent: Agent-computer interfaces enable automated software engineering, 2024{\natexlab{a}}.
\newblock URL \url{https://arxiv.org/abs/2405.15793}.

\bibitem[Yang et~al.(2024{\natexlab{b}})Yang, Peng, Wang, and Zhang]{Yang2024LLMbasedMS}
Yingxuan Yang, Qiuying Peng, Jun Wang, and Weinan Zhang.
\newblock Llm-based multi-agent systems: Techniques and business perspectives.
\newblock 2024{\natexlab{b}}.
\newblock URL \url{https://api.semanticscholar.org/CorpusID:274165614}.

\bibitem[Yang et~al.(2025{\natexlab{a}})Yang, Chai, Song, Qi, Wen, Li, Liao, Hu, Lin, Chang, Liu, Wen, Yu, and Zhang]{yang2025surveyaiagentprotocols}
Yingxuan Yang, Huacan Chai, Yuanyi Song, Siyuan Qi, Muning Wen, Ning Li, Junwei Liao, Haoyi Hu, Jianghao Lin, Gaowei Chang, Weiwen Liu, Ying Wen, Yong Yu, and Weinan Zhang.
\newblock A survey of ai agent protocols, 2025{\natexlab{a}}.
\newblock URL \url{https://arxiv.org/abs/2504.16736}.

\bibitem[Yang et~al.(2025{\natexlab{b}})Yang, Ma, Huang, Chai, Gong, Geng, Zhou, Wen, Fang, Chen, Gu, Jin, Spanos, Yang, Abbeel, Song, Zhang, and Wang]{yang2025agenticwebweavingweb}
Yingxuan Yang, Mulei Ma, Yuxuan Huang, Huacan Chai, Chenyu Gong, Haoran Geng, Yuanjian Zhou, Ying Wen, Meng Fang, Muhao Chen, Shangding Gu, Ming Jin, Costas Spanos, Yang Yang, Pieter Abbeel, Dawn Song, Weinan Zhang, and Jun Wang.
\newblock Agentic web: Weaving the next web with ai agents, 2025{\natexlab{b}}.
\newblock URL \url{https://arxiv.org/abs/2507.21206}.

\bibitem[Yao et~al.(2023)Yao, Zhao, Yu, Du, Shafran, Narasimhan, and Cao]{yao2023reactsynergizingreasoningacting}
Shunyu Yao, Jeffrey Zhao, Dian Yu, Nan Du, Izhak Shafran, Karthik Narasimhan, and Yuan Cao.
\newblock React: Synergizing reasoning and acting in language models, 2023.
\newblock URL \url{https://arxiv.org/abs/2210.03629}.

\bibitem[Zan et~al.(2023)Zan, Chen, Zhang, Lu, Wu, Guan, Wang, and Lou]{zan2023NL2code}
Daoguang Zan, Bei Chen, Fengji Zhang, Dianjie Lu, Bingchao Wu, Bei Guan, Yongji Wang, and Jian-Guang Lou.
\newblock Large language models meet nl2code: A survey, 2023.
\newblock URL \url{https://arxiv.org/abs/2212.09420}.

\bibitem[Zhang et~al.(2025{\natexlab{a}})Zhang, Yue, Sun, Wan, Yu, Fang, Wang, Chen, and Cheng]{zhang2025gdesigner}
Guibin Zhang, Yanwei Yue, Xiangguo Sun, Guancheng Wan, Miao Yu, Junfeng Fang, Kun Wang, Tianlong Chen, and Dawei Cheng.
\newblock G-designer: Architecting multi-agent communication topologies via graph neural networks, 2025{\natexlab{a}}.
\newblock URL \url{https://arxiv.org/abs/2410.11782}.

\bibitem[Zhang et~al.(2025{\natexlab{b}})Zhang, Li, Wang, Chen, Zhang, Ye, Feng, Wang, Wang, Wang, Xu, Bai, Ouyang, and Hu]{zhang2025avengers}
Yiqun Zhang, Hao Li, Chenxu Wang, Linyao Chen, Qiaosheng Zhang, Peng Ye, Shi Feng, Daling Wang, Zhen Wang, Xinrun Wang, Jia Xu, Lei Bai, Wanli Ouyang, and Shuyue Hu.
\newblock The avengers: A simple recipe for uniting smaller language models to challenge proprietary giants, 2025{\natexlab{b}}.
\newblock URL \url{https://arxiv.org/abs/2505.19797}.

\end{thebibliography}
\end{document}